\documentclass[letterpaper, 10 pt, conference]{ieeeconf}
\IEEEoverridecommandlockouts
\usepackage{graphics}
\usepackage{epsfig}
\usepackage{stfloats}
\usepackage{color, colortbl}
\definecolor{LightCyan}{rgb}{0.88,1,1}
\newcommand*{\myalign}[2]{\multicolumn{1}{#1}{#2}}

\title{\LARGE \bf
Multi-Resolution Elevation Mapping and Safe Landing Site Detection with Applications to Planetary Rotorcraft 
}

\author{Pascal Schoppmann$^{1,2}$, Pedro F. Proen\c{c}a$^{1}$, Jeff Delaune$^{1}$, Michael Pantic$^{2}$, Timo Hinzmann$^{2}$,\\ Larry Matthies$^{1}$, Roland Siegwart$^{2}$ and Roland Brockers$^{1}$

\thanks{$^{1}$ Jet Propulsion Laboratory / California Institute of Technology, Pasadena, CA, USA.
{\tt\small \{pproenca, jeff.h.delaune, lhm, roland.brockers\} @jpl.nasa.gov}}
\thanks{$^{2}$ Autonomous Systems Lab, ETH Zurich, Switzerland. 
{\tt\small \{pascscho, michael.pantic, timo.hinzmann, rsiegwart\} @ethz.ch}}
}

\begin{document}

\maketitle
\thispagestyle{empty}
\pagestyle{empty}

\begin{abstract}
	In this paper, we propose a resource-efficient approach to provide an autonomous UAV with an on-board perception method to detect safe, hazard-free landing sites during flights over complex 3D terrain. We aggregate 3D measurements acquired from a sequence of monocular images by a Structure-from-Motion approach into a local, robot-centric, multi-resolution elevation map of the overflown terrain, which fuses depth measurements according to their lateral surface resolution (pixel-footprint) in a probabilistic framework based on the concept of dynamic Level of Detail. Map aggregation only requires depth maps and the associated poses, which are obtained from an on-board Visual Odometry algorithm. An efficient landing site detection method then exploits the features of the underlying multi-resolution map to detect safe landing sites based on slope, roughness, and quality of the reconstructed terrain surface. The evaluation of the performance of the mapping and landing site detection modules are analyzed independently and jointly in simulated and real-world experiments in order to establish the efficacy of the proposed approach.
\end{abstract}

\section{Introduction}
Recent developments by NASA indicate that unmanned aerial vehicles (UAVs) could make a significant contribution to future Mars exploration, since autonomous UAVs enable science missions beyond the reach of orbiters, landed spacecrafts or rovers \cite{mars}\cite{marsSH2020}. \textit{Ingenuity}, NASA's Mars Helicopter, which recently landed on Mars, demonstrated the first powered flight on another planet. While \textit{Ingenuity} assumes flat and level terrain during flight \cite{ingenuity_VIO}, future planetary rotorcrafts need the ability to fly fully autonomously over complex 3D terrain, which is one of the reasons why NASA is interested in safe landing site detection in previously unknown and unstructured terrain.

A critical part in the development of autonomous rotorcrafts is to ensure their safety. Therefore, it is essential that the vehicle has a mean of perceiving and understanding its surroundings as accurately and certain as possible at all times. In this way, hazardous objects can be detected and collisions avoided. To land in a safe place at the end of a mission or in case of an emergency, such as rapid battery degradation or sensor failure, future planetary rotorcrafts have to be able to detect safe landing sites in previously unknown and challenging terrain autonomously and without human intervention. 

The landing site detection system on-board is constrained to size, weight and power. Since weight has a considerable influence on energy consumption and thus on flight duration, a lightweight and energy-efficient sensor and computing architecture is required. In addition, the algorithms used for mapping and landing site detection must be efficient enough to be applied in real time on an embedded system with limited resources. In this paper, we utilise a single down-facing navigation camera as a light-weight sensor option.

\begin{figure} []
    \begin{center}
        \includegraphics[width=0.9\columnwidth]{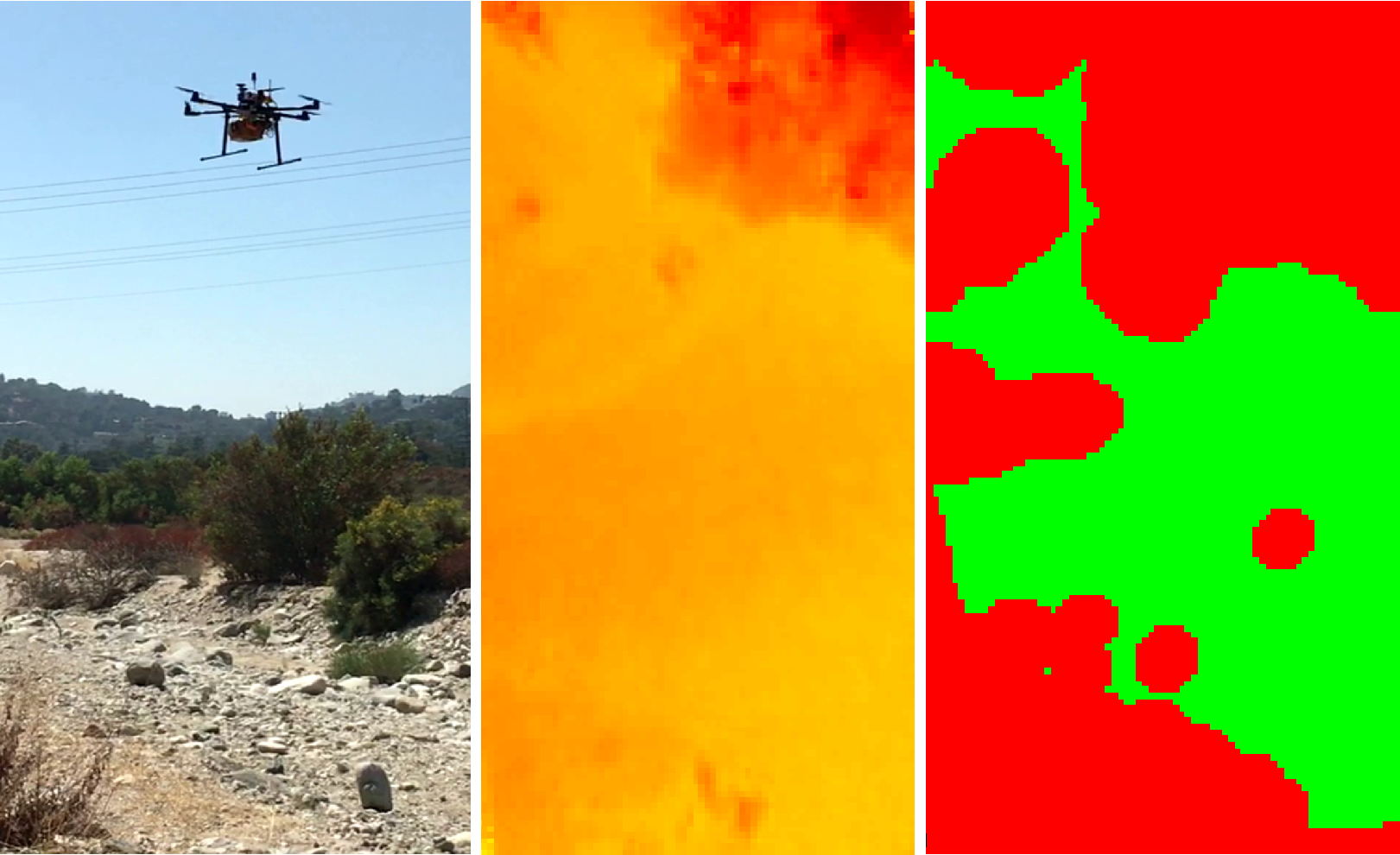}
        \caption{UAV flight in outdoor environment. Middle: Top-down view of color-coded elevation map (warmer colors are closer to the camera); Right: Associated landing site map (Green: safe landing sites, Red: hazards).}
        \label{drone}
    \end{center}
    \vspace{-13pt}
\end{figure}

The presented work proposes a system that enables UAVs to map their local environment and find safe landing sites. In order to enable the system to be used in remote and previously unknown areas or in an emergency, we limit ourselves to using only the computing capacity available on-board. To achieve this goal, we use a Structure-from-Motion (SfM) algorithm that is coupled with an on-board state estimator to generate dense and metric depth maps. However, since these depth maps are noisy and incomplete, the depth maps are temporally fused into a consistent model of the environment in order to cope with individual noise and outliers. As surface representation, a 2D robot-centric, multi-resolution digital elevation map is used, that maintains a hierarchy of approximations of the environment at multiple scales, sufficient for many autonomous maneuvers and safe landing site detection in outdoor environments. Finally, we present a landing site detection approach, which efficiently exploits the multi-resolution structure of the underlying map by applying a coarse-to-fine search approach based on different cost functions, such as slope, roughness and uncertainty.\\

The contribution of this work is a landing site detection framework that is executed continuously on a small embedded processor on-board an autonomous rotorcraft. Summarising, our contributions are the following:

\begin{itemize}
	\item A local, robot-centric, 2.5D multi-resolution digital elevation grid map that aggregates vision-based 3D measurements seamlessly based on their corresponding pixel footprint.
	\item A lightweight map fusion algorithm that incorporates depth measurements according to their lateral surface resolution (pixel-footprint) using Kalman filter.
	\item An efficient landing site detector that extracts safe landing sites based on slope, roughness and map quality, exploiting the multi-resolution structure.
\end{itemize}

In addition, we present a qualitative and quantitative evaluation on synthetic and real data sets, which reveals the performance of the proposed framework.\\

\section{Related Work}
Autonomous UAV landing has been widely discussed in the UAV literature. Several vision-based approaches have been proposed to provide landing sites with known artificial markings, which can be detected on individual images \cite{helipad}\cite{movingPlatform}.

Approaches to detect safe landing sites from monocular image sequences are presented in \cite{homography}\cite{ homography_Brockers}. Using a homography assumption, an incremental model is constructed that only includes horizontal, flat surfaces. Therefore, the map can not be used for obstacle avoidance. \cite{rooftop}\cite{rooftop_Brockers} propose a vision-based landing site detection framework to estimate a planar rooftop landing area by deploying a homography strategy. However, the use of a homography in complex 3D terrain is not feasible. For the purpose of spacecraft landing, NASA has developed an autonomous landing hazard avoidance system using a Lidar for 3D perception of landing hazards \cite{NASAlanding}\cite{marsLidar}. \cite{lidarLandingMars}\cite{lidarLanding} use range sensors for large-scale UAV landing. However, given weight and size constraints, range sensors are not suitable for a weight restricted UAV. The Mars 2020 mission deploys LVS, a lander vision system to detect landing hazards \cite{LVS}. Given an on-board map with predetermined hazard locations, LVS uses a monocular camera to estimate the spacecraft's position during descent and triggers an avoidance maneuver if necessary. Unfortunately, landing hazards for UAVs are much smaller than for landers, and maps with the resolution required to detect UAV landing hazards off-line are not available. Additionally, previously acquired maps cannot annotate dynamic landing hazards (e.g. cars or persons on Earth).

Similar to our work, \cite{hazardTerrain} describes vision-based autonomous landing in unknown, 3D terrain. Also using an SfM approach, a dense point cloud is calculated and projected onto a digital elevation model (DEM). The DEM is then evaluated for roughness and slope to segment safe and unsafe landing sites. While this work determines the landing site entirely on a single DEM, we continuously fuse depth measurements into a local elevation map to temporally improve the map representation in the present of 3D depth errors. Most recent works fuse a stream of depth maps from an SfM approach into an 2.5D elevation map \cite{elevationMap}\cite{contLanding}\cite{nightLanding}. These works apply a simple arithmetic on the elevation map, by using a local neighborhood operator to check if the surrounding cells have a similar height to detect flat and obstacle free regions. While in \cite{elevationMap}\cite{nightLanding} the map resolution is tied to the pixel footprint, \cite{contLanding} switches the resolution of the elevation map depending on the flight altitude, but does not store multiple resolutions and therefore information is lost during re-sampling.

In \cite{multiRes} a fixed surface map is adapted to incorporate depth maps with multiple resolutions. Depth maps are directly fused into a multi-resolution triangular mesh based on a regular grid. The proposed algorithm is computationally demanding and requires a high-end GPU. While this is not feasible for on-board processing on a small embedded processor, we follow the idea of a multi-resolution map to fuse image-based 3D measurements based on the local pixel resolution, but implement a different, grid-based surface representation.

\section{System Overview}
\begin{figure} []
	\begin{center}
		\includegraphics[width=1\columnwidth]{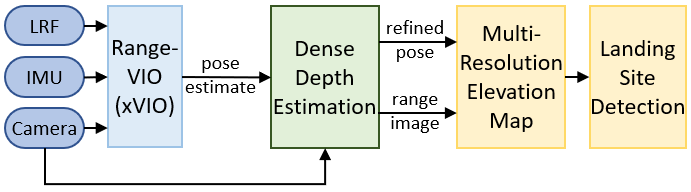}
		\caption{Overview of the main modules of the complete landing system, running on-board the UAV. \label{systemOverview}}
	\end{center}
	\vspace{-5pt}
\end{figure}

Figure \ref{systemOverview} gives an overview of the proposed processing pipeline on-board the UAV. A range visual-inertial odometry algorithm (xVIO) estimates the current pose of the UAV using a down-facing camera, an IMU, and a laser range finder (LRF). xVIO combines measurements from these three sensors in a tightly coupled approach \cite{xVIO} to overcome traditional weaknesses of VIO such as scale unobservability in the absence of inertial excitation, e.g. during critical maneuvers such as constant-velocity traverses or hovering with no motion. The state estimator is a standalone module for robustness purposes, and for computational efficiency.

While pose estimates are accurate enough for controlling the UAV, they are not sufficient for dense 3D reconstruction. Therefore, we deploy a camera pose refinement step \cite{dense_3d}, using the outputs from the estimator as pose priors. Based on baseline and feature-related constraints, two images are selected and a conventional real-time stereo algorithm \cite{stereo} is deployed to calculate a dense stereo disparity image. Finally, by triangulating the disparity image, a range image containing the 3D positions of each pixel in the world frame is achieved. Range images are then used to incrementally build a robot-centric, multi-resolution digital elevation map adapted for landing applications. Whenever a new range image is available, the corresponding Level of Detail (LoD) is calculated for every 3D point in the image based on the footprint of the corresponding pixel, and a Bayesian update step is performed to update affected cells. This has the advantage, that the map representation can incorporate measurements naturally based on their lateral resolution with no need for resampling, as will be explained in Section~\ref{multiResMap}.

Whenever the elevation map is updated, the landing site detector evaluates the map in several steps exploiting the multi-resolution structure and using various cost functions, as will be described in Section \ref{lsd} resulting in a binary map indicating if a map cell is a feasible landing site or not.

\section{Multi-Resolution Elevation Mapping}
\label{multiResMap}
The SfM system described in the previous Section can efficiently create depth maps and thus enable a UAV to perceive its 3D environment. However, since depth maps tend to be noisy and incomplete, due to non-Lambertian surfaces, occlusions, or textureless regions, their direct use for safe landing site detection is limited. Hence, it is necessary to temporally fuse them into a consistent model of the UAV's environment. Furthermore, when a UAV overflies a surface at different altitudes, depth maps with different resolutions are estimated and need to be incorporated into a single map.

\subsection{Multi-scale Surface Representation}
\begin{figure} []
	\begin{center}
		\includegraphics[width=0.8\columnwidth]{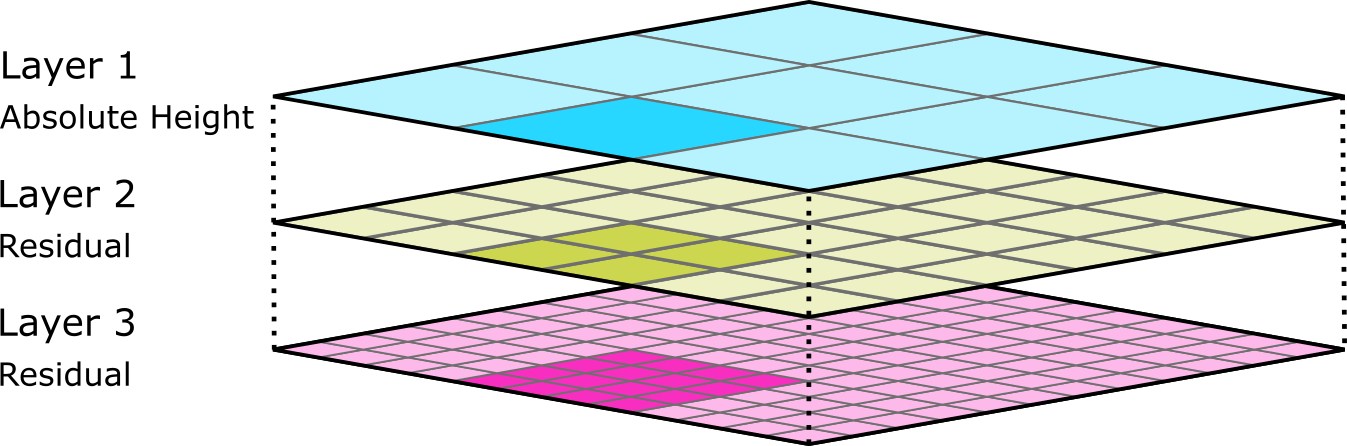}
		\caption{Multi-resolution map structure. Blue: base layer storing the absolute height of the aggregated measurements; Yellow and Red: successive layers carry residuals. Slightly darker coloured squares in each layer cover the same ground area. \label{map_structure}}
	\end{center}
	\vspace{-5pt}
\end{figure}

We fuse range measurements into a multi-resolution elevation map that is based on a Laplacian pyramid decomposition representing the observed environment with multiple scales in a single, consistent model. Figure~\ref{map_structure} illustrates the structure of the map, where each layer contains a regular sampled 2D grid map with predefined, fixed topology. Subsequent layers are sub-sampled by a factor of two, allowing to incorporate new measurements probabilistically in a coarse-to-fine manner.

Inspired by the Laplace pyramid by Burt-Adelson \cite{pyramid}, the layers within the pyramid contain different frequencies of the surface structure, where finer resolved layers contain higher frequencies not captured by the coarser layers. An advantage of the Laplacian pyramid is that individual layers can be assumed to be independent, which simplifies fusion and enables recursive estimation. The map does not store the environment at multiple resolutions as widely used in computer graphics, but uses an implicit representation, where the coarsest layer contains the aggregated height values of all measurement within the footprint of a coarse cell. Subsequent layers only store frequency components called residuals calculated by

\begin{equation}
	r_{k+1} = z - h_k = z - \left(h_1 + \sum_{n=2}^{k} r_n \right)
\end{equation}

where $z$ is the measured elevation and $h_k$ the aggregated height, $h_1$ respectively is the height stored in the coarsest layer (base layer, or layer 1).

\begin{figure} []
	\begin{center}
		\includegraphics[width=0.99\columnwidth]{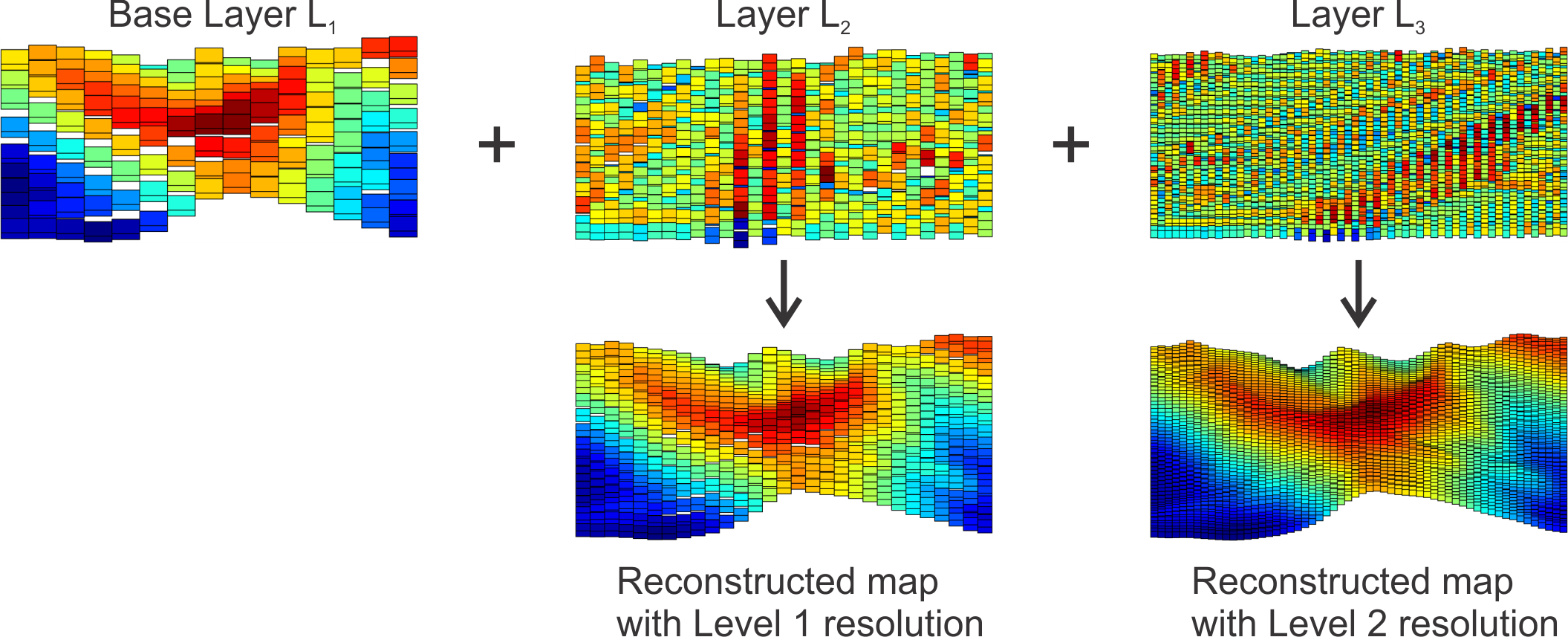}
		\caption{Surface reconstruction from multi-resolution map. The higher the layer in the pyramid added, the more details appear. \label{reconstruction}}
	\end{center}
	\vspace{-5pt}
\end{figure}

Reconstruction of the surface is straight forward by simply adding the different layers of the pyramid (Figure~\ref{reconstruction}). Given a position $(x,y)$ on the surface, the extracted height is calculated by

\begin{equation}
	h_l(x,y) = h_1(x,y) + \sum_{n=2}^{l} r_n(x,y)
\end{equation}

where $h_1(x,y)$ is the absolute height of the base layer with respect to the UAVs initial position while $r_n(x,y)$ are the residuals at position $(x,y)$ and pyramid level $n$.\\

The layout described enables cell retrieval in their corresponding coarser or finer grid maps efficiently and in a simple manner. Cells at any scale are denoted by their integer coordinates $x_l$ at the finest resolution. If $d$ is the maximum depth of the pyramid and $l$ the desired layer, the scaling factor is given by $s = 2^{d-l}$ with $l \leq d$. Hence, index coordinates to access the multi-resolution grid at any resolution $l$ can be obtained by simple arithmetic as:

\begin{equation}
	x_l = \bigg\lfloor \frac{x_d}{s} \bigg\rfloor
\end{equation}

where $x_d$ represents the index of the cell at the finest resolution. The memory overhead of the Burt-Adelson pyramid scheme is 4/3 compared to a single layer map \cite{pyramid}.

\subsection{Dynamic Level of Detail}
When flying over 3D terrain or when the camera is mounted at an oblique angle, the pixel footprint of a measurement varies within a depth map. Therefore, it is not practical to work with a single global map resolution, since a single resolved map typically results in aliasing artifacts when the selected resolution is too high or details disappear if the resolution is too small. To cope with measurements with varying resolutions, we apply a dynamic level of detail concept which is inspired by computer graphics methods to adapt the complexity of an object to the expected on-screen pixel resolution. Applying the inverse process, we incorporate a measurement only up to the level of the Laplacian pyramid with the corresponding pixel resolution. We first compare the pixel footprint of a measurement i, representing the area of a pixel in the range map projected on the surface,

\begin{equation}
	\label{eq_LOD}
	p_{xi} = \frac{2(z_a - z_i)\cdot\tan(\frac{FOV_x}{2})}{n_x}
\end{equation}

with the resolutions represented in the map levels. Here $z_a$ is the current altitude of the UAV, $z_i$ the elevation of the measurement, $FOV$ the field of view of the camera and $n_x$ the size of one image row. Therefore, the parameter controlling the LoD is the distance of a point to the camera. Finally, the level with the next lower resolution to the measurement resolution is selected as the target level. By only incorporating the measurements up to the desired resolved layer, we can minimize aliasing artifacts.

\subsection{Map Update}
Given the pyramid structure in which the coarsest layer contains absolute height values and the subsequent layers the residuals, it is possible to directly estimate the coefficients of the Laplacian pyramid. Each depth map is processed as it arrives to incrementally update the map in a coarse-to-fine manner. Updates are first applied to the base level, then for each finer level. Residuals are calculated and fused until the required level of detail has been reached, repeated recursively for each individual 3D measurement. Individual map cells are updated by a Kalman update step, using a measurement variance directly derived from the expected maximum stereo disparity error of 0.25~px, neglecting pose uncertainty. Given the height measurement $h_i$ which corresponds to the z-axis of the world coordinates, and the height variance $\sigma_i$ which is calculated given a baseline from the 3D reconstruction process, the recursive Kalman update step to update the height estimate $h_p$ and variance $\sigma_p$ of the corresponding cell is formulated as \cite{kalman};

\begin{equation}
	h_p \leftarrow \frac{h_p\sigma_i^2 + h_i\sigma_p^2}{\sigma_p^2 + \sigma_i^2}
\end{equation}

\begin{equation}
	\sigma_p^2 \leftarrow \frac{\sigma_p^2 \cdot \sigma_i^2}{\sigma_p^2 + \sigma_i^2}
\end{equation}

Note, that the same measurement uncertainty $\sigma_i$ is applied in all levels, since the predicted height is assumed to be constant among layers.

\subsection{Map Movement}
\begin{figure}
	\begin{center}
		\includegraphics[width=0.85\columnwidth]{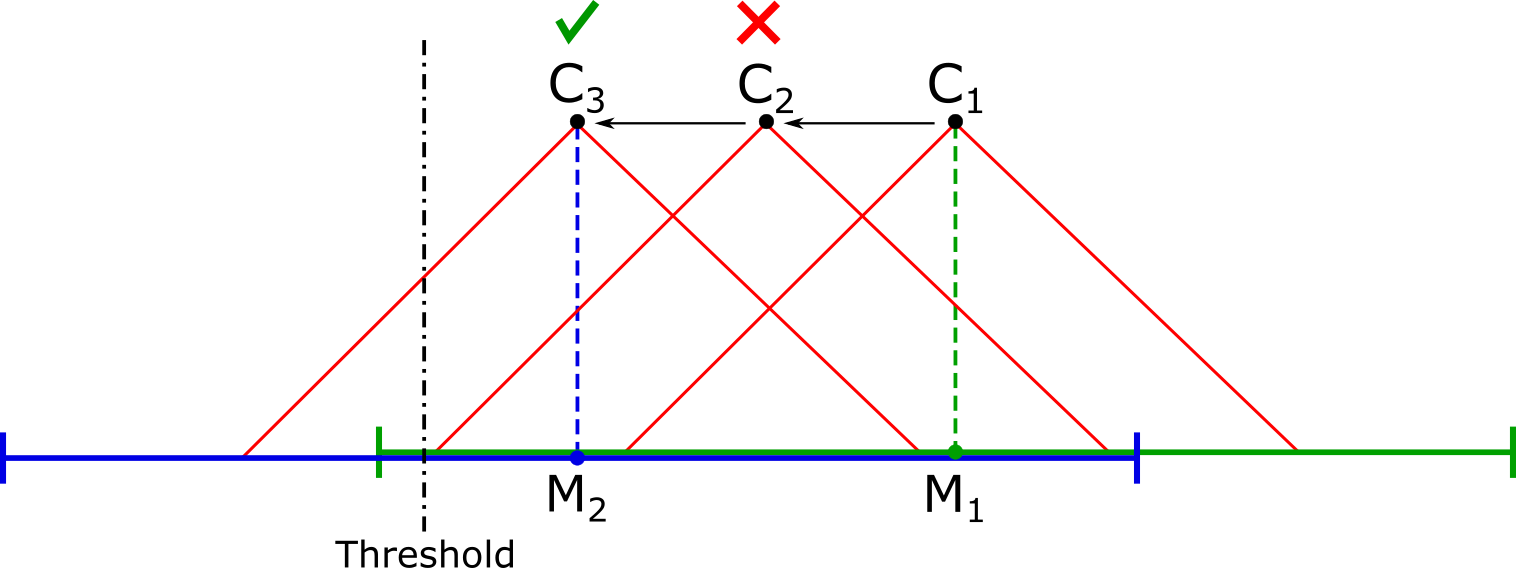}
		\caption{Scheme of the map movement logic. When the UAV moves from position C$_1$ to C$_2$ the map is not moved since the new measurements still fall into the map area. At the third position C$_3$ the measurements lie outside the boundary of the map, and the map is moved from position M$_1$ to M$_2$. \label{map_move}}
	\end{center}
	\vspace{-6pt}
\end{figure}

The map with fixed size is implemented using a two-dimensional rolling buffer. Therefore, the map requires constant memory and can be moved efficiently by shifting indices and resetting cells that move out of the map area. Beside being memory efficient, a rolling map is non-destructive. In order to prevent loss of data while moving back and forth, the map is only moved when a measurement at a new position falls outside the map boundaries as depicted in Figure \ref{map_move}. We make the map pseudo-robot-centric and locate the map directly underneath the UAV to deal with drift in the pose estimate.

\subsection{Discussion}
A regular sampled elevation map is a straight forward 3D surface representation for UAVs with limited computational capabilities. An obvious drawback of the regular grid is that cells borders do not correspond to natural features of the surface and the measurement resolution does not need to correspond to an existing layer resolution. Being aware of this drawback, we find that the proposed representation is sufficient for the task of autonomous UAV landing site detection. Artifacts can occur in the areas within the map where the resolution changes between neighboring cells. However, since this is only the case in non-flat areas of the map, we can ignore those artifacts for the purpose of landing site detection.

\section{Landing Site Detection}
\label{lsd}
Robust and safe landing site detection is essential to mitigate the risk of crash landings. Since emergency landings might be required at any time during flight a landing site detection algorithm needs to be efficient enough to run on-board in near real-time.

\subsection{Landing Requirements}
The requirements for adequate landing sites are defined by the landing gear of the UAV. Thus, we define a safe landing site to have a local neighborhood of certain radius sufficiently large for descending and landing in which the following criteria are fulfilled:

\begin{itemize}
	\item \textit{Slope}: The inclination of the surface is below a maximum threshold.
	\vspace{4pt}
	\item \textit{Roughness}: The surface within the landing area is sufficiently flat, respectively free of obstacles.
	\vspace{4pt}
    \item \textit{Confidence}: The landing site is detected with sufficient confidence. If the confidence is low, e.g. because only a few measurement have covered an area on the map, no conclusion can be drawn about the safety of a landing site.\\
\end{itemize}

A safe landing area consists of a \textit{keep-out-zone} defined by the size of the UAV and an additional \textit{safety margin}, which is introduced to alleviate quantization effects of the regular sampled grid as will be discussed in Section \ref{experiments}.

\begin{figure}
	\begin{center}
		\includegraphics[width=0.9\columnwidth]{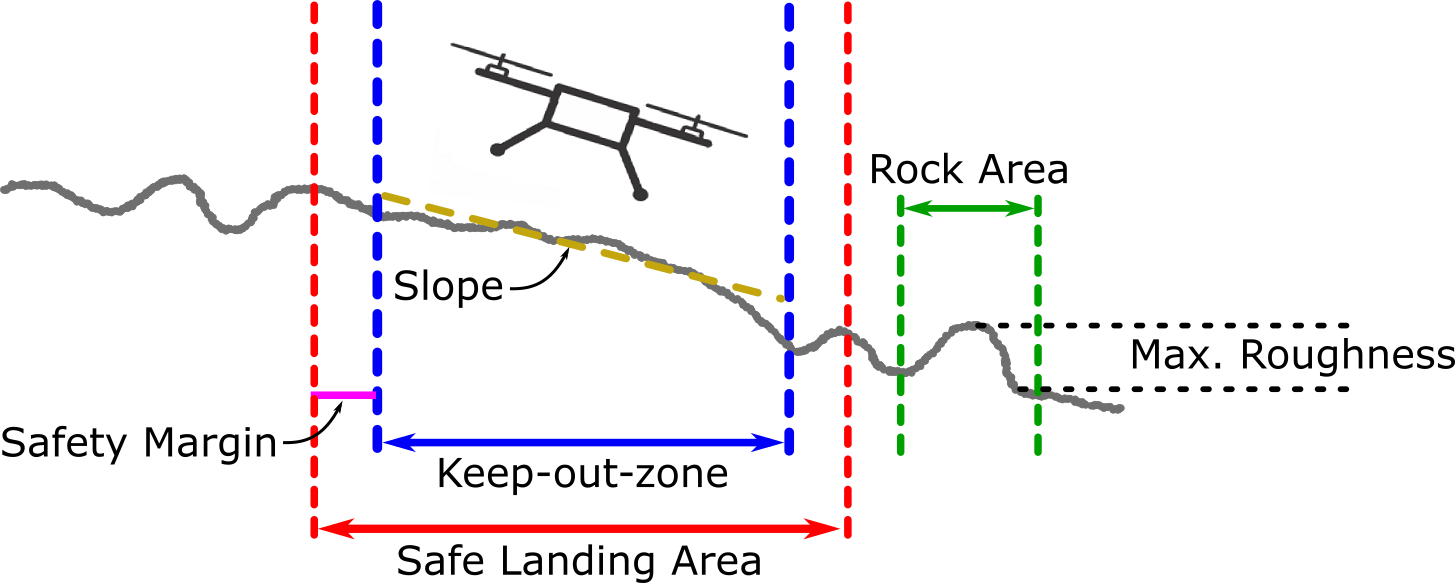}
		\caption{Requirements for a landing site. Within the \textit{Safe Landing Area}, slope, and roughness must fulfill the requirements set by the landing gear. \label{landing}}
	\end{center}
	\vspace{-5pt}
\end{figure}

\subsection{Evaluation Criteria}
The landing site detector analyzes the elevation map in several stages, considering two basic assumptions:

\begin{itemize}
	\item \textit{Asm. 1}: The slope of the aggregated layers is similar to the slope of the coarsest layer.
	\vspace{5pt}
	\item \textit{Asm. 2}: If Asm. 1 does not hold true, the roughness criteria is violated.\\
\end{itemize}

Therefore, the detector first evaluates the coarsest layer for slope, and - if successful - then performs a hazard analysis in the finer layers. Further, it is assumed that areas which are unsafe in coarser layers are also not safe in finer layers. Therefore, if an area is declared as unsafe in a coarse layer it is no longer evaluated in subsequent layers, saving computation time. However, since the size of detectable hazards decreases with the resolution increase of finer layers, a coarse-to-fine evaluation of the roughness criteria is still required. The roughness evaluation consists of two local neighborhood operator. First, the roughness criteria is verified using a smaller \textit{Rock Area} (Figure \ref{landing}) accounting for abrupt changes around hazards. Second, the complete \textit{Safe Landing Area} is verified to be sufficiently planar.

\begin{figure*} [!b]
\centering
 	\begin{tabular}{@{}c@{\hspace{5pt}}c@{\hspace{5pt}}c@{\hspace{5pt}}c@{}}%cccc}
    \includegraphics[width=0.25\textwidth]{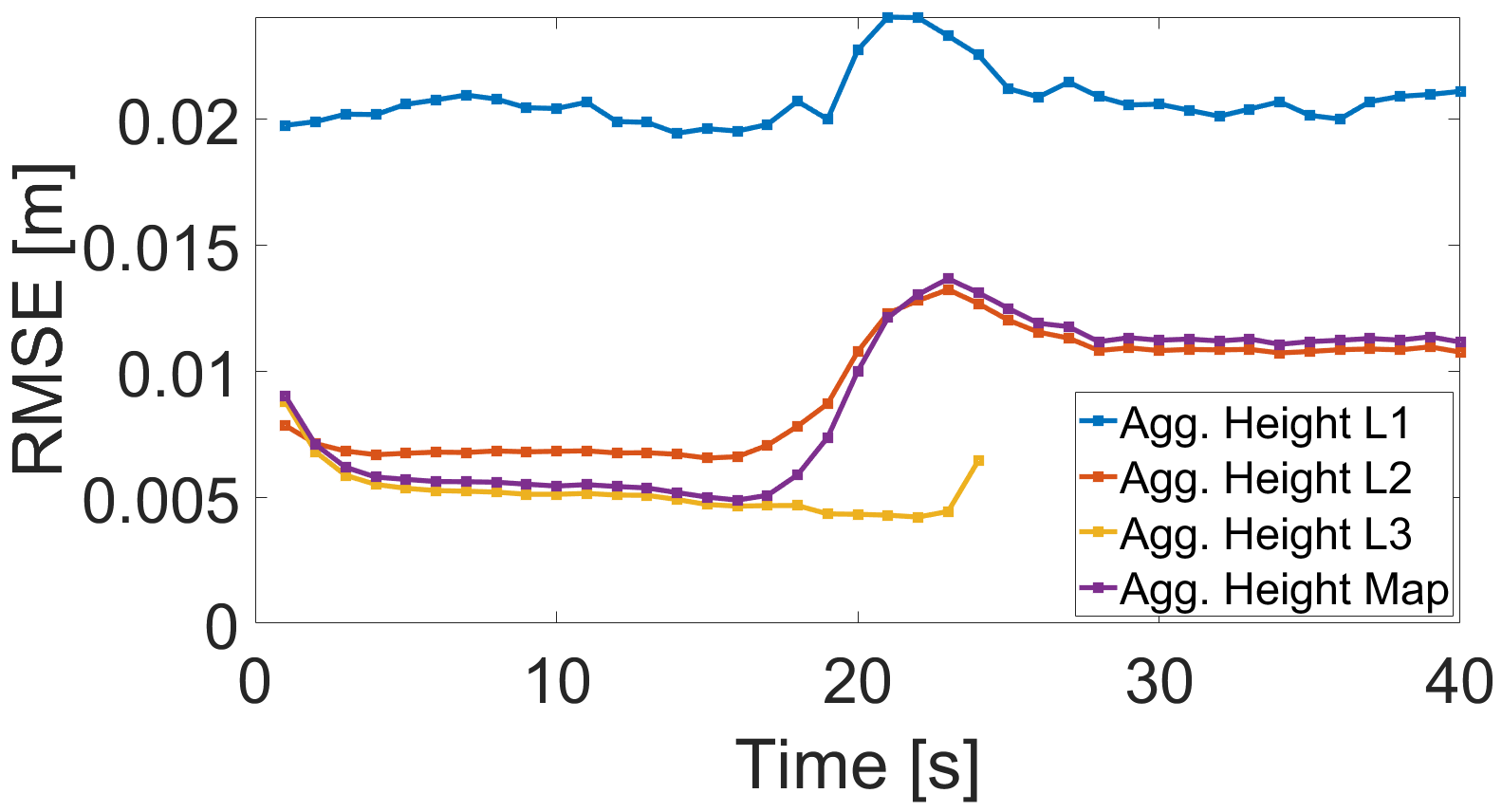} &
    \includegraphics[width=0.25\textwidth]{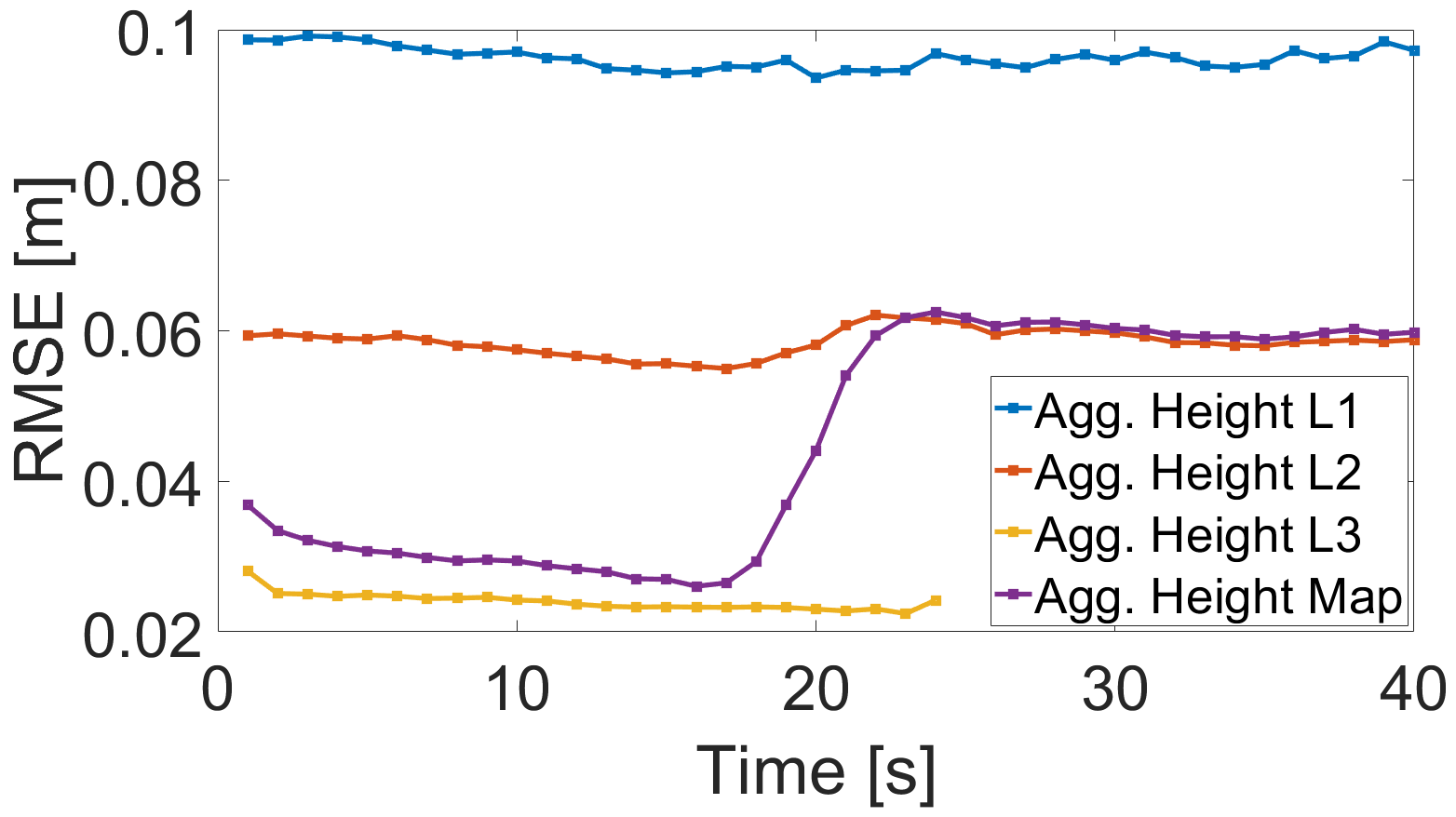} & 
    \includegraphics[width=0.25\textwidth]{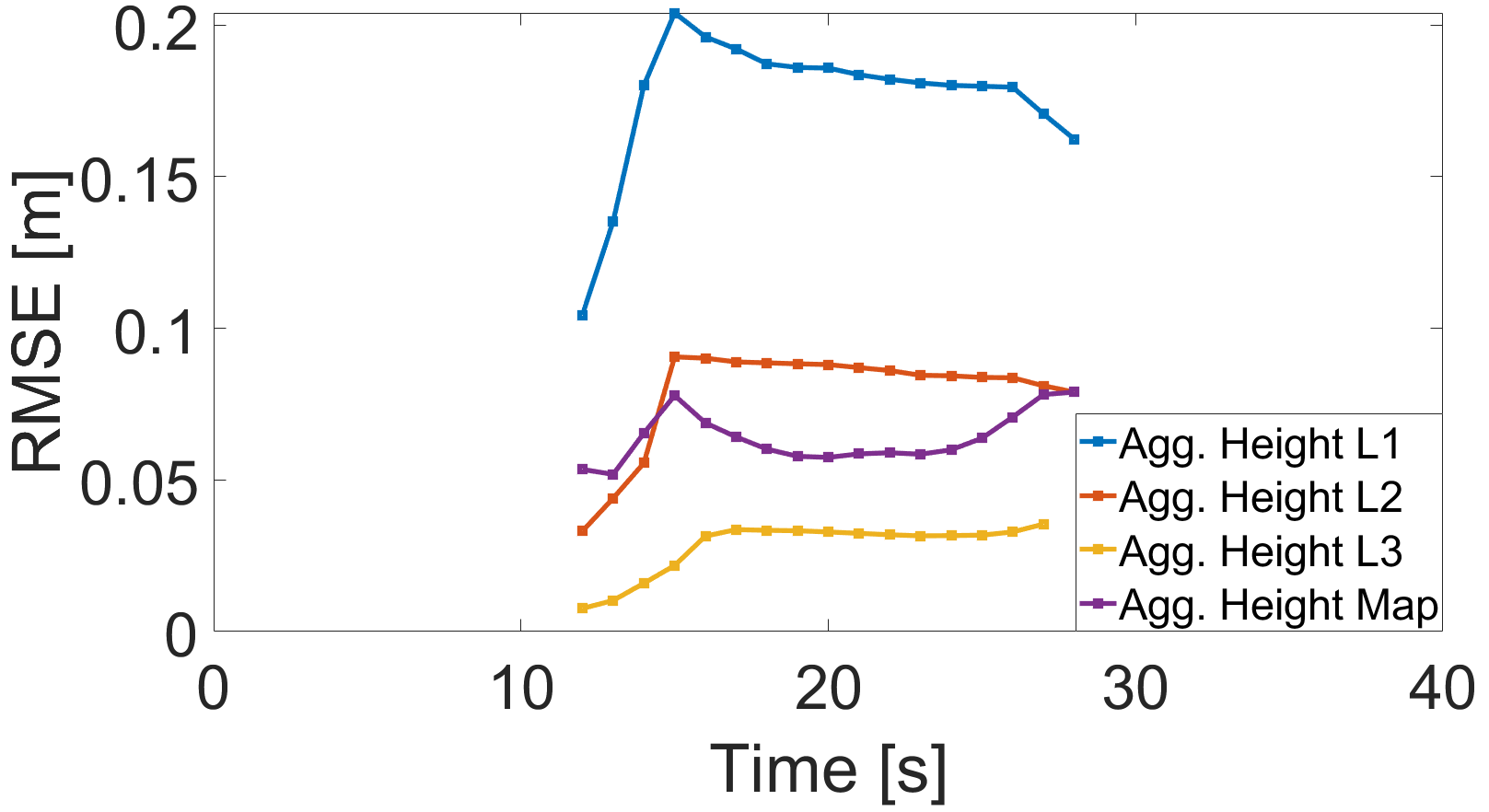} &
 	\raisebox{0.2\height}{\includegraphics[width=0.205\textwidth]{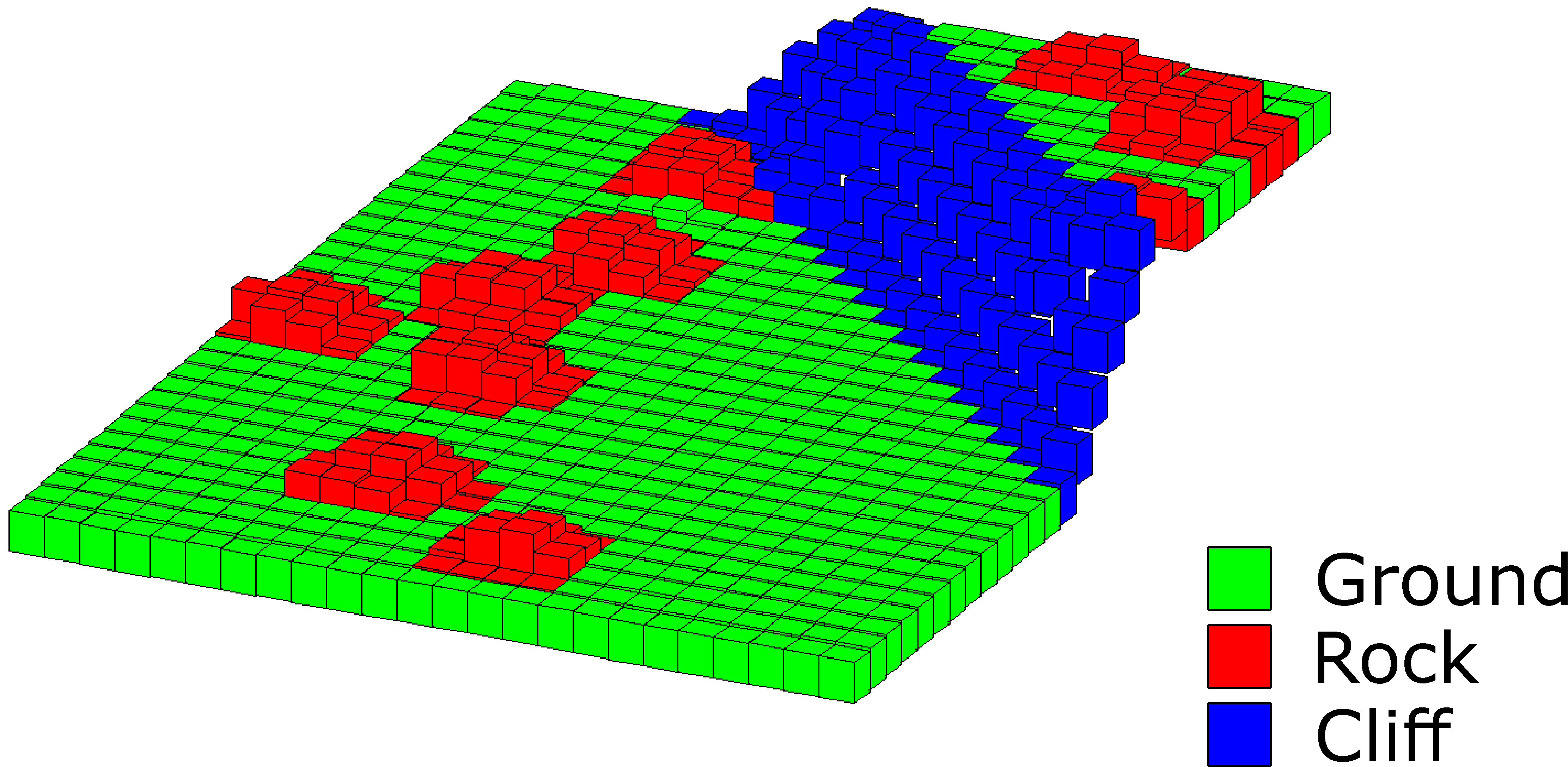}} \\
    (a) & (b) & (c) & (d)
	\end{tabular}
  \caption{Root Mean Square Error (RMSE) of the reconstructed map calculated for map cells that contain (a) flat ground, (b) rocks with a diameter of 30~cm, and (c) cliffs according to the terrain segmentation illustrated in (d). Flight altitude: 5~m AGL (before cliff), 10~m AGL (after cliff). Map: 3~layers, 8~cm resolution at highest level; terrain slope: 5$^\circ$, cliff height: 5~m.}
  \label{fig:map_precision_eval}
\end{figure*}

The landing site detector result is a binary map which annotates if a map cell is a valid landing site or not. Finally, by applying a distance transform, locations with a maximum distance to any obstacles can be selected and collected in a list of landing site candidates to be considered by the UAV on-board autonomy.

\section{Experiments}
\label{experiments}
We run various experiments on synthetic and real flight data sets to evaluate the performance of our proposed framework.

\subsection{Mapping Experiments}\label{mapping_experiments}
We evaluate the map quality with simulated terrain consisting of a nominally flat but rough Brownian surface \cite{Brownian} with a pre-defined inclination, overlaid with rocks, represented as half spheres of a defined size, and a cliff (see Figure \ref{fig:map_precision_eval}d). Rocks are randomly distributed, given a desired rock abundance (20$\%$ of the total surface). 3D point clouds were generated by simulating a stereo camera with a fixed altitude depending baseline corresponding to $\sim$80\% image overlap with added Gaussian noise corresponding to a 0.25 pixel disparity error (3~sigma). For this experiment the UAV starts at an altitude of 5~m above ground level (AGL) in an elevated area and flies over a 5~m cliff resulting in an altitude of 10~m~AGL.

During flight the map is updated continuously and for each update we perform a map evaluation for each of the three terrain types (Figure \ref{fig:map_precision_eval}). 

As expected, the map is most accurate, when measurements are incorporated in the highest resolution layers. At $t = 12$~s, the cliff starts to appear in the images. Since the distance to the terrain and thus the pixel footprint increases while flying over the cliff, measurements from below the cliff are no longer incorporated into the finest layer, which results in an increased total map Root Mean Square Error (RMSE) converging to the RMSE of the layer 2 map reconstruction. At $t = 28$~s, the cliff is out of view, and therefore layer 3 does not contain any data for the rest of the flight. By comparing the three evaluations, we can see that the mapping module can accurately reconstruct the ground plane, while the RMSE for rocks and cliffs is higher. Since we are only interested in safe landing sites located on flat ground, this is acceptable for our approach.

\subsection{Landing Experiments}
\begin{figure} [t]
	\begin{center}
		\includegraphics[width=0.95\columnwidth]{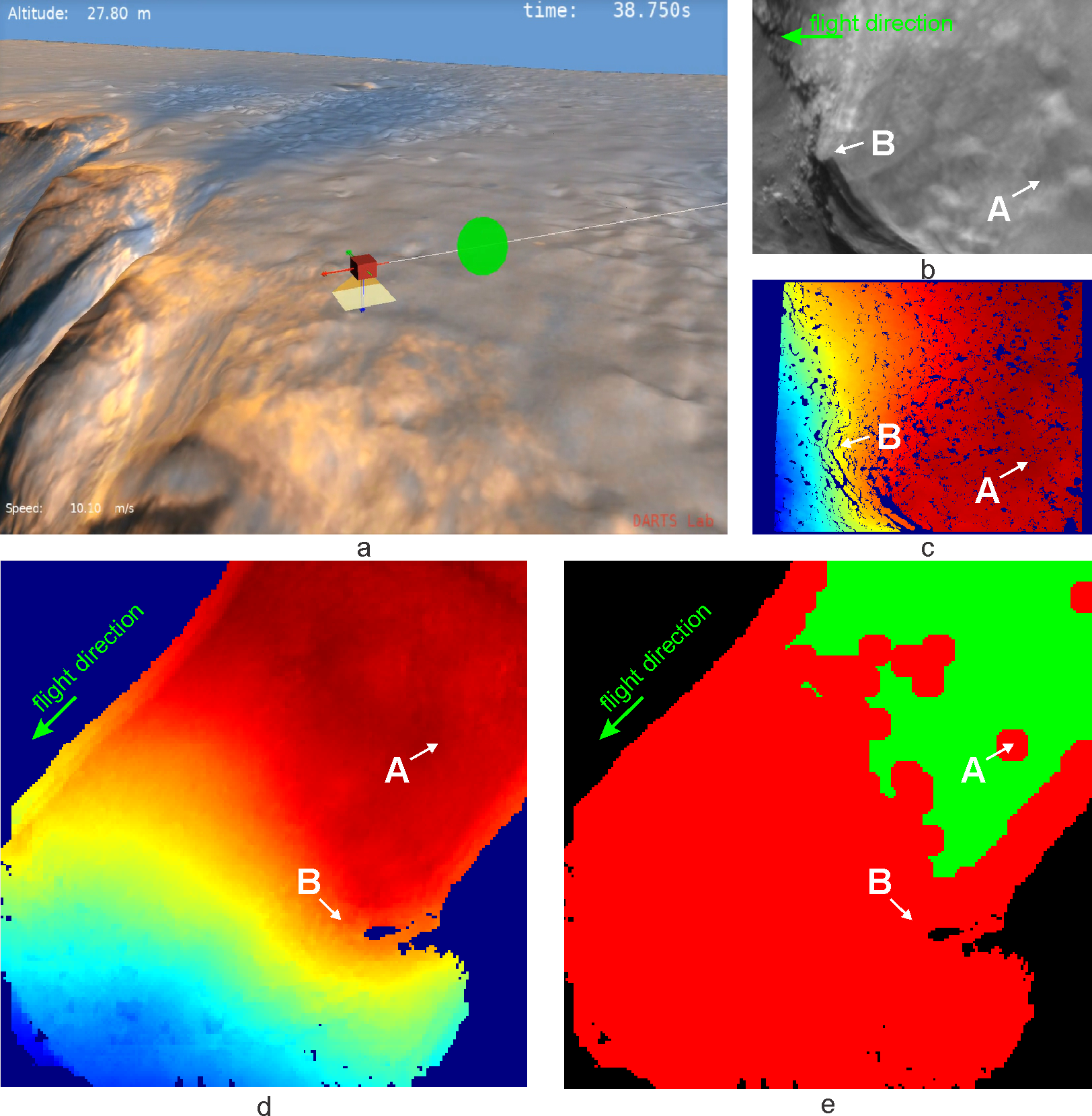}
		\caption{(a) Simulated UAV flight over Mars Victoria crater rim; b) Current reference view (rectified left image); c) Height map generated from stereo disparity map (warmer colors are closer to camera); d) Aggregated elevation map (top-down view). Note, that map is rotated $\sim$45$^\circ$; e) Landing site map (green: safe landing site, red: landing hazard, black: no data). A and B label selected landing hazards for illustration. Flight altitude: 20~m~AGL. Map: 3~layers, 10~cm highest res.; safety area radius: 1.0~m; slope threshold: 20$^\circ$. 
		\label{syntheticData}}
	\end{center}
	\vspace{-15pt}
\end{figure}

Figure \ref{syntheticData} illustrates a simulated UAV flight using the JPL DARTS simulator \cite{darts}. A sequence of artificial images was used to calculate dense depth maps with our SfM module, which were temporally fused within the map representation. As can be seen, landing sites are correctly detected on the flat terrain outside the crater, but not on the terrain below the rim that violates the slope constraint.

\begin{figure} [t]
	\begin{center}
		\vspace{15pt}
		\includegraphics[width=0.78\columnwidth]{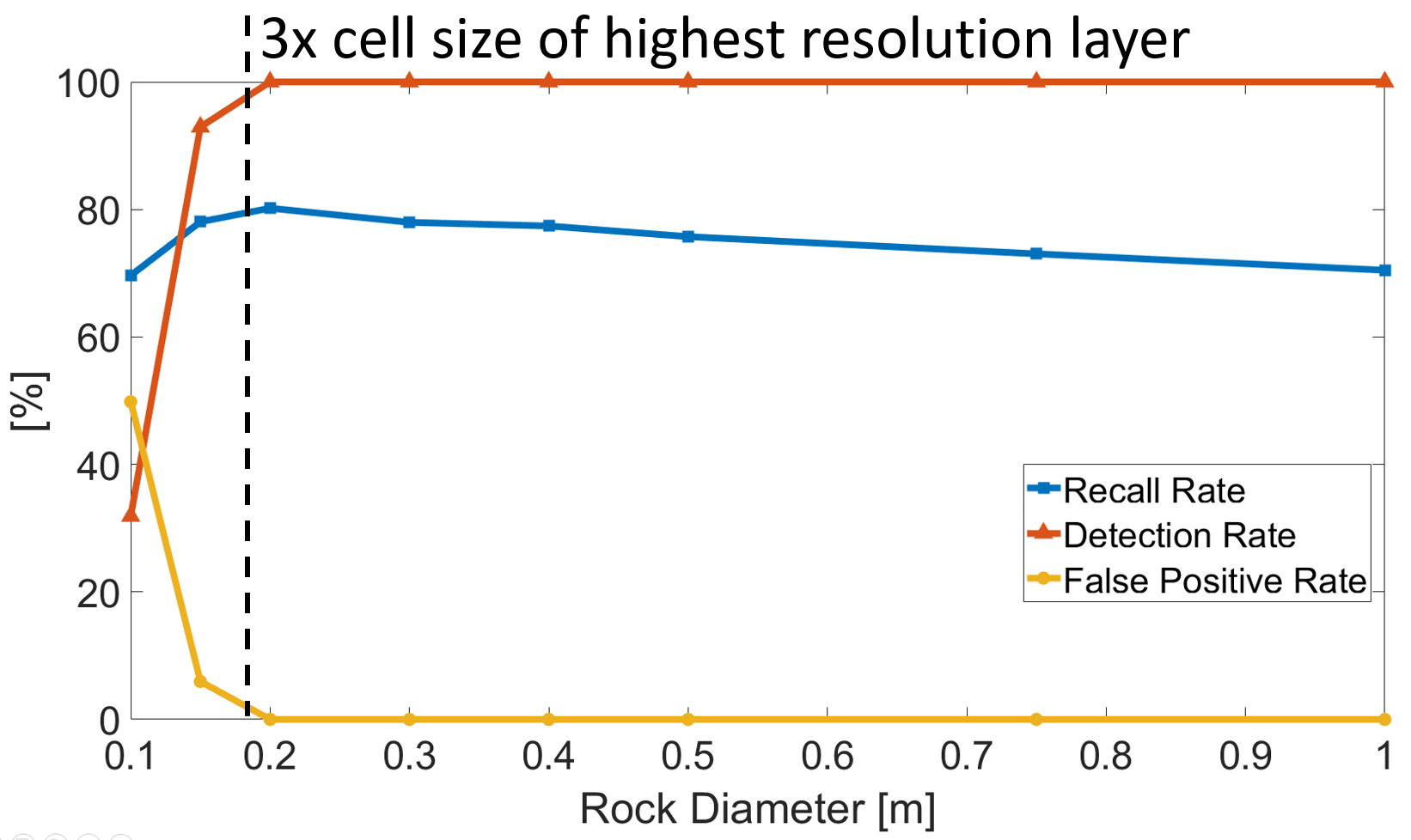}
		\caption{Landing site detection evaluation. By using a safe landing area that includes a safety margin next to the keep-out-zone, the False-Positive-rate can be minimized. Flight altitude: 5-6~m; terrain slope: 5$^\circ$. Map: 3 layers, highest resolution: 6~cm. Safe landing area radius: 0.6~m; keep-out zone radius: 0.5~m; max. slope: 10$^\circ$.}
    \label{rockSizeEval}
	\end{center}
\end{figure}

In order to determine the performance limits of the landing site detector, we use the same artificially generated surface as described in section \ref{mapping_experiments} to verify the ability of the landing site detector to detect small landing hazards. 16 flights were executed over terrain consisting of a slightly sloped ground plane (5$^\circ$) with randomly distributed rocks with a fixed rock coverage area of 20$\%$ and rock sizes as shown in Figure~\ref{rockSizeEval} and Table~\ref{landingTable}. For each flight, rocks of a fixed size were randomly distributed on a ground plane, and 50 individual frames were evaluated, leading to a total average rock count of 2000 rocks per rock size. Map areas within one \textit{safe landing area} radius of a non-assigned map cell (map border) were excluded from the evaluation since the detector annotates these areas as hazardous by design.

\begin{figure} [t]
	\begin{center}
		\includegraphics[width=0.95\columnwidth]{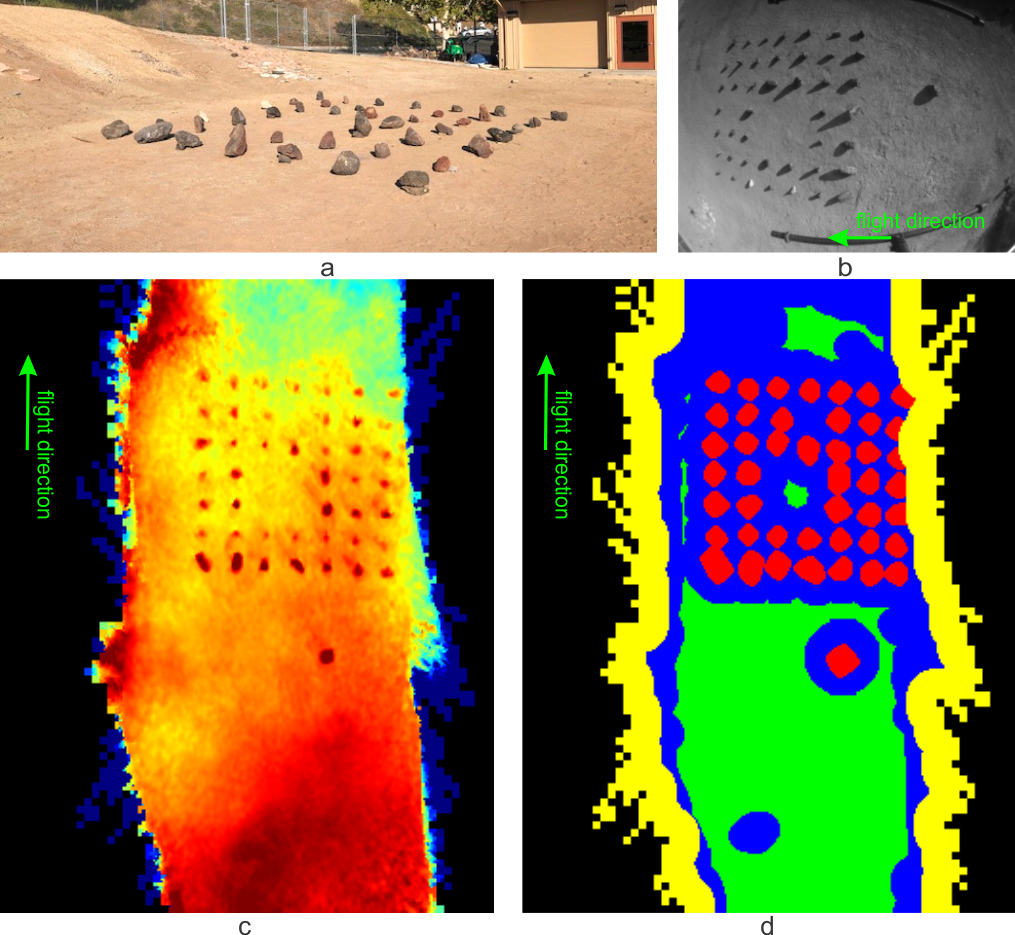}
		\caption{UAV flight over a rock field. a) 7x7 rock field; b) Raw camera image; c) Aggregated elevation map; d) Evaluated landing site map based on annotated rocks (green: safe landing site, blue: false landing hazard, yellow: border region assumed to be hazardous, not evaluated, red: correctly detected hazards). Flight altitude: 5 m. Map: 3 layers. Safety area radius: 0.5~m, keep-out-zone radius used for evaluation: 0.3~m, slope threshold: 10$^\circ$, rock area radius: 0.5~m, max. roughness: 0.1~m.}
    \label{MY4_flight}
	\end{center}
	\vspace{-15pt}
\end{figure}
\begin{table}[b]
    \centering
    \caption{Recall, Detection, and False Positives rates for different sizes of landing hazards. M(light blue): with safety margin.}
    \label{landingTable}
    \setlength{\tabcolsep}{0.5em}
    \begin{tabular}{l|c|c|c|c|c|c}
        \hline
        Rock Diameter {[}m{]}  & 0.1 & 0.2 & 0.3 & 0.4 & 0.5 & 1.0 \\ \hline\hline
        Recall Rate {[}\%{]} & \myalign{r|}{68.8} & \myalign{r|}{87.5} & \myalign{r|}{85.6} & \myalign{r|}{84.8} & \myalign{r|}{82.6} & \myalign{r}{77.3}  \\ \hline
        Detection Rate {[}\%{]} & \myalign{r|}{31.8} & \myalign{r|}{92.9} & \myalign{r|}{100.0} & \myalign{r|}{100.0} & \myalign{r|}{100.0} & \myalign{r}{100.0} \\ \hline
        False Pos. Rate {[}\%{]} & \myalign{r|}{60.2} & \myalign{r|}{0.5} & \myalign{r|}{0.4} & \myalign{r|}{0.2} & \myalign{r|}{0.1} & \myalign{r}{0.1} \\ \hline
        \rowcolor{LightCyan}
        Recall Rate M {[}\%{]} & \myalign{r|}{69.6} & \myalign{r|}{80.2} & \myalign{r|}{78.0} & \myalign{r|}{77.4} & \myalign{r|}{75.7} & \myalign{r}{70.5} \\ \hline
        \rowcolor{LightCyan}
        Detection Rate M {[}\%{]} & \myalign{r|}{33.4} & \myalign{r|}{93.5} & \myalign{r|}{100.0} & \myalign{r|}{100.0} & \myalign{r|}{100.0} & \myalign{r}{100.0} \\ \hline
        \rowcolor{LightCyan}
        False Pos. Rate M {[}\%{]} & \myalign{r|}{44.1} & \myalign{r|}{0.003} & \myalign{r|}{0.0} & \myalign{r|}{0.0} & \myalign{r|}{0.0} & \myalign{r}{0.0} \\ \hline
    \end{tabular}
\end{table}

The detector is capable of resolving rock sizes larger than 3 times the highest cell resolution as can be seen in the high \textit{detection rate} - the number of rocks detected in relation to the number of rocks that were visible during flight (a rock counts as detected when all cells which contain a part of a rock are correctly segmented as hazardous). Additionally, we are able to reduce the false-positive-rate - the number of falsely detected safe landing sites - to zero by the introduction of an additional safety margin (Figure~\ref{landing}). This however comes at the cost of a reduced \textit{recall rate} - the number of correctly segmented cells divided by the total number of cells in a map, which is acceptable, since a conservative result that minimises the risk of a crash landing rather than finding all possible landing sites is preferred.

\begin{table}[t]
    \centering
    \caption{Rock detection rates for different map cell sizes for the UAV data set [\%].}
    \label{MY4_eval_table}
    \setlength{\tabcolsep}{0.5em}
    \begin{tabular}{l|c|c|c|c}
        \hline
        Rock Height {[}m{]} & 0.14-0.22 & 0.22-0.30 & 0.30-0.38 & 0.38-0.46 \\ \hline\hline
        Rock Diameter* {[}m{]} & 0.28-0.44 & 0.44-0.60 & 0.60-0.76 & 0.76-0.92 \\ \hline\hline
        3~cm Cell Size {[}\%{]} & \myalign{r|}{100.0} & \myalign{r|}{100.0} & \myalign{r|}{100.0} & \myalign{r}{100.0} \\ \hline
        6~cm Cell Size {[}\%{]} & \myalign{r|}{100.0} & \myalign{r|}{100.0} & \myalign{r|}{100.0} & \myalign{r}{100.0} \\ \hline
        12~cm Cell Size {[}\%{]} & \myalign{r|}{98.4} & \myalign{r|}{100.0} & \myalign{r|}{100.0} & \myalign{r}{100.0} \\ \hline
        20~cm Cell Size {[}\%{]} & \myalign{r|}{59.3} & \myalign{r|}{89.1} & \myalign{r|}{100.0} & \myalign{r}{100.0} \\ \hline\hline
        Number of Rocks ** & 25 & 14 & 4 & 2 \\ \hline
    \end{tabular}
    \begin{tabular}{l}
        * Note, that rock diameter is approximated using a half sphere model \\
        *** Rock height distribution of rocks in the rock field (Figure \ref{MY4_flight} a)
    \end{tabular}
\end{table}

\begin{figure} [t]
	\begin{center}
		\includegraphics[width=1.0\columnwidth]{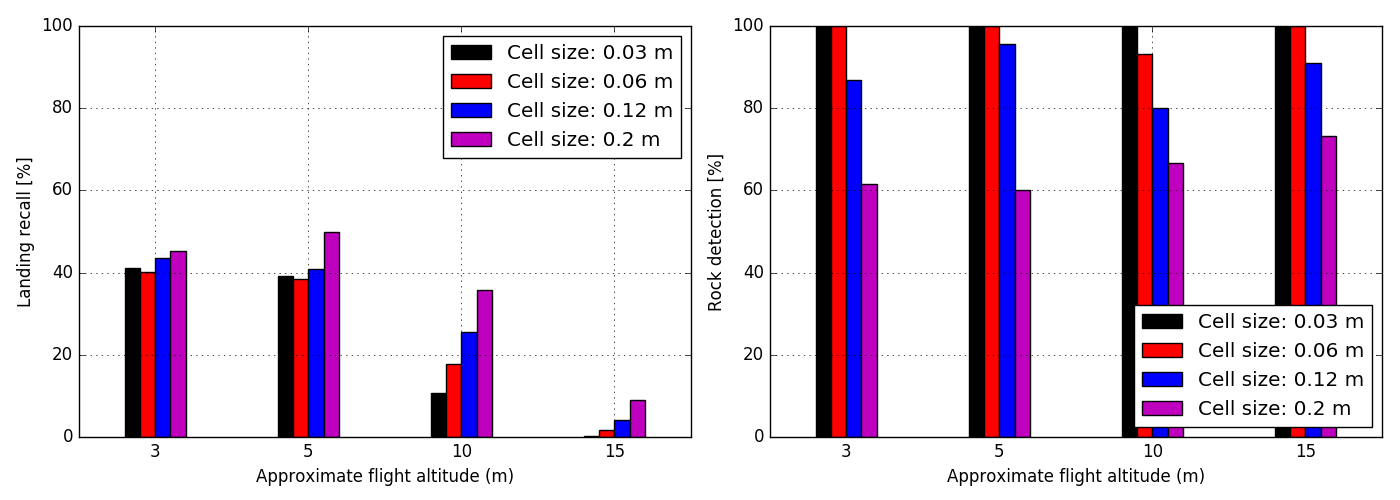}
		\caption{Landing recall (left) and rock detection rate (right) over different map cell sizes for multiple UAV flights over the rock field shown in Figure \ref{MY4_flight} at different altitudes.}
    \label{MY4_eval_different_altitudes}
	\end{center}
	\vspace{-15pt}
\end{figure}

To verify the simulation results, we conducted flights with a UAV (TurboAce Infinity-6, equipped with a Snapdragon Flight Pro) over a rock field (Figure~\ref{MY4_flight}a) consisting of rocks with heights between 14~cm and 46~cm. Rocks were selected, such that they can be approximated as half spheres, where the rock height corresponds to half the rock diameter. Images were captured with a nadir pointed camera with VGA resolution and 110$^\circ$ horizontal field of view at a flight altitude of approximately 5~m and a velocity of 1~m/s. Landing site detection and evaluation were performed offline. 

Similar to the simulation results in Figure \ref{rockSizeEval}, the rock detection rate starts decreasing if rocks have an approximated diameter below 3x the cell size in the highest resolution layer (Table~\ref{MY4_eval_table}). 

To evaluate landing site performance as a function of the distance to the overflown surface, we additionally conducted flights at different altitudes. As can be seen in Figure \ref{MY4_eval_different_altitudes}, the recall rate decreases at higher altitudes. This is caused by increased noise on stereo range measurements which scales with distance, and fewer updates in individual map cells in the highest resolution layer. The result is that the map surface is no longer smooth and the detection of small rocks becomes more difficult as expected.

\begin{figure} [t]
	\begin{center}
		\includegraphics[width=0.95\columnwidth]{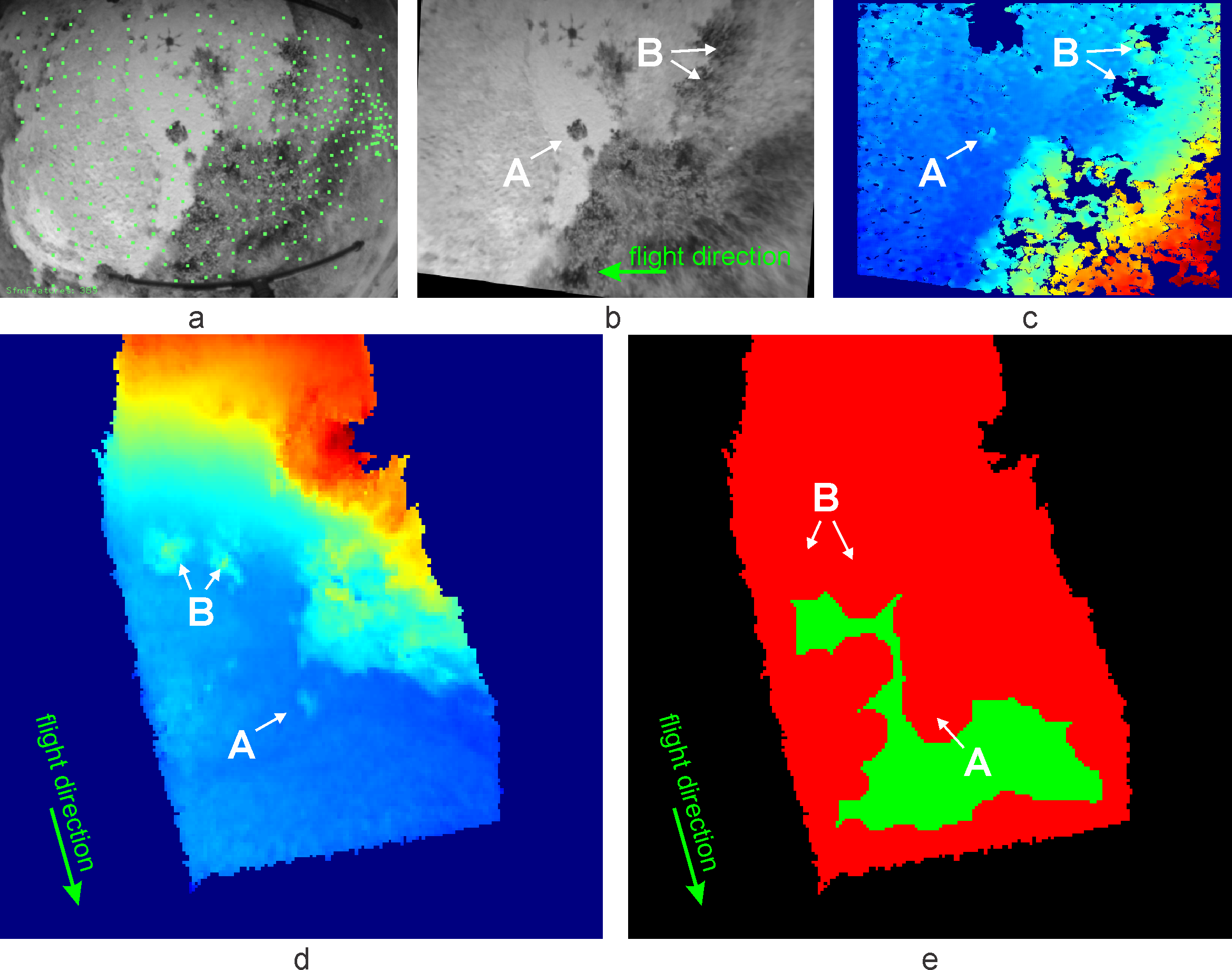}
		\caption{Landing site detection with UAV flight data: a) Raw camera image with overlaid features that are tracked by the state estimator; b) Reference view (rectified); c) Height map; d) Aggregated elevation map; e) Landing site map (green: safe landing site, red: landing hazard; black: no data). A and B label selected landing hazards for illustration. Note, that the map is rotated $\sim$105$^\circ$. Flight altitude: 8~m. Map: 3~layers, highest resolution: 10~cm. Safety area radius: 0.5~m, max.~slope: 10$^\circ$.}
    \label{arroyolanding}
	\end{center}
	\vspace{-15pt}
\end{figure}

Figure \ref{arroyolanding} illustrates landing site detection results with UAV flight data over arid terrain. Our approach is able to resolve individual, small landing hazards and detect a 15$^\circ$ slope at the right side of the camera image.

\subsection{Runtime Evaluation}
For run time measurements, we conducted a test flight in simulation at an altitude of 20~m~AGL at a speed of approximately 2~m/s and measured run-times on an Intel Xeon E-2286M, 2.4~GHz processor, with both, the mapping and landing site detection module, running on a single core. A VGA-resolution depth map is processed by the mapping module in approximately 18.2~ms ($\sigma = 1.69$~ms) to update a map of 16~m x 16~m size with 3~layers and a maximal resolution of 8~cm per cell -  corresponding to a total number of 52,500 cell updates using 0.4~MB memory. The landing site detection module evaluated this map in approximately 20.3~ms ($\sigma =  ~1.95$~ms). Note, that the execution time however depends greatly on the altitude of the UAV and the structure of the observed scene. We expect execution times to scale to an embedded processor to meet a targeted detection frame rate of approximately 1 Hz.

\section{Conclusions and Future Work}
We presented a framework to autonomously detect safe landing sites on-board a future planetary rotorcraft with limited size, weight and power resources. The proposed elevation mapping approach is capable of incrementally modelling a surface by processing vision-based 3D measurements with multiple resolutions in a dynamic Level of Detail approach. While being memory and computationally efficient, the multi-resolution map refines a coarse terrain approximation with local high-resolution information. The presented landing site detector exploits the multi-resolution structure of the representation of the environment and allows a fast and efficient detection of safe landing site. We clearly see the multi-resolution mapping approach advantageous in 3D terrain with various elevations, or for cameras mounted with an oblique angle, where our approach allows to create a highly detailed map close to the vehicle and a coarse map further away or in  areas with less dense measurements. The framework was tested on various simulated and real environments, validating the feasibility and robustness of our vision-based methods.

\section{Acknowledgment}
The research was carried out at the Jet Propulsion Laboratory, California Institute of Technology, under a contract with the National Aeronautics and Space Administration (80NM0018D0004) and was in part supported by the National Center of Competence in Research (NCCR) on Digital Fabrication through the Swiss National Science Foundation.


\begin{thebibliography}{99}

\bibitem{mars} J. Balaram, T. Canham, C. Duncan, H. Grip, W. Johnson, J. Maki, A. Quon, R. Stern, and D. Zhu, ``Mars Helicopter Technology Demonstrator'', AIAA SciTech Forum, 2018

\bibitem{marsSH2020} W. Johnson, S. Withrow, L. Young, W. Koning, W. Kuang, C. Malpica, J. Balaram and T. Tzanetos, ``Mars Science Helicopter Conceptual Design", AIAA Ascend 2020

\bibitem{ingenuity_VIO}D. Bayard, D. Conway, R. Brockers, J. Delaune, L. Matthies, H. Grip, G. B. Merewether, T. L. Brown and A. M. San Martin, ``Vision-Based Navigation for the NASA Mars Helicopter", AIAA Scitech Forum, 2019

\bibitem{helipad} S. Saripalli, J. F. Montgomery, and G. S. Sukhatme, ``Vision-based autonomous landing of an unmanned aerial vehicle,” in \textit{IEEE International Conference on Robotics and Automation (ICRA)}, vol. 3, 2002

\bibitem{movingPlatform} D. Falanga, A. Zanchettin, A. Simovic, J. Delmerico, and D. Scaramuzza, ``Vision-based autonomous quadrotor landing on a moving platform”, in \textit{Proceedings of the IEEE International Symposium on Safety, Security and Rescue Robotics}, Shanghai, China, 2017

\bibitem{homography} S. Bosch, S. Lacroix and F. Caballero, ``Autonomous Detection of Safe Landing Areas for an UAV from Monocular Images," in \textit{IEEE/RSJ International Conference on Intelligent Robots and Systems}, Beijing, 2006

\bibitem{homography_Brockers} R. Brockers, P. Bouffard, J. Ma, L. Matthies, and C. Tomlin, ``Autonomous landing and ingress of micro-air-vehicles in urban environments based on monocular vision", in \textit{SPIE Defense, Security and Sensing}, 2011

\bibitem{rooftop} V. Desaraju, M. Humenberger, N. Michael, R. Brockers, S.Weiss and L. Matthies, ``Vision-based Landing Site Evaluation and Trajectory Generation Toward Rooftop Landing", in \textit{Autonomous Robots}, vol 39, no. 3, pp. 445-463, 2015

\bibitem{rooftop_Brockers} R. Brockers, M. Hummenberger, S. Weiss and L. Matthies, ``Towards Autonomous Navigation of Miniature UAV," in \textit{IEEE Conference on Computer Vision and Pattern Recognition Workshops}, pp. 645-651, 2014

\bibitem{NASAlanding} N. Trawny, A. Huertas, M. Luna, C. Villalpando, K. Martin, J. Carson and A. Johnson, "Flight testing a real-time hazard detection system for safe lunar landing on the rocket-powered mopheus vehicle", in \textit{Proc. AIAA SciTech Conference}, 2015

\bibitem{marsLidar} M. Luna, E. Almeida, G. Spiers, C. Villalpando, A. Johnson, and N. Trawny, ``Evaluation of the Simple Safe Site Selection (S4) Hazard Detection Algorithm using Helicopter Field Test Data", in \textit{AIAA Guidance, Navigation, and Control Conference}, 2017

\bibitem{lidarLandingMars} A. Johnson, A. Klumpp, J. Collier, and A .Wolf, ``Lidar-based hazard avoidance for safe landing on mars," AIAA Jour. in \textit{Guidance, Control and Dynamics}, vol. 25, no. 5, 2002

\bibitem{lidarLanding} S. Scherer, L. Chamberlain, and S. Singh, ``Autonomous landing at unprepared sites by a full-scale helicopter", in \textit{Journal of Robotics and Autonomous Systems}, 2012

\bibitem{LVS} A. Johnson, N. Villaume, C. Umsted, A. Kourchians, D. Sternberg, N. Trawny, Y. Cheng, E. Giepel, and J. Montgomery, ``The Mars 2020 lander vision system field test", in \textit{Proc. AAS Guidance Navigation and Control Conference} (AAS-20-105), 2020

\bibitem{hazardTerrain} A. Johnson, J. Montgomery, and L. Matthies, ``Vision guided landing of an autonomous helicopter in hazardous terrain", in \textit{IEEE Intl. Conf. on Robotics and Automation (ICRA)}, 2005

\bibitem{elevationMap} P. Fankhauser, M. Bloesch, C. Gehring, M. Hutter, and R. Siegwart, ``Robot-centric elevation mapping with uncertainty estimates," in \textit{International Conference on Climbing and Walking Robots (CLAWAR)}, 2014

\bibitem{contLanding} C. Forster, M. Faessler, F. Fontana, M. Werlberger and D. Scaramuzza, ``Continuous on-board monocular-vision based elevation mapping applied to autonomous landing of micro aerial vehicles", in \textit{IEEE International Conference on Robotics and Automation (ICRA)} , 2015

\bibitem{nightLanding} S. Daftry, M. Das, J. Delaune, C. Sorice, R. Hewitt, S. Reddy, D. Lytle, E. Gu and L. Matthies, ``Robust Vision-based Autonomous Navigation, Mapping and Landing for MAVs at Night", in \textit{International Symposium on Experimental Robotics (ISER)}, 2018

\bibitem{multiRes} J. Zienkiewicz, A. Tsiotsios, A. Davison and S. Leutenegger, ``Monocular, Real-Time Surface Reconstruction Using Dynamic Level of Detail," in \textit{Fourth International Conference on 3D Vision (3DV)}, pp. 37-46, 2016

\bibitem{xVIO} J. Delaune, R. Brockers, D. S. Bayard, H. Dor, R. Hewitt, J. Sawoniewicz, G. Kubiak, T. Tzanetos, L. Matthies and J. Balaram, ``Extended Navigation Capabilities for a Future Mars Science Helicopter Concept", in \textit{IEEE Aerospace Conference}, 2020

\bibitem{dense_3d} M. Domnik, P. Proenca, J. Delaune, J. Thiem and R. Brockers, ``Dense 3D-Reconstruction from Monocular Image Sequences for Computationally Constrained UAS", in \textit{IEEE Winter Conference on Applications of Computer Vision (WACV)}, 2021

\bibitem{stereo} S. B. Goldberg, M. Maimone and L. Matthies, ``Stereo vision and rover navigation software for planetary exploration", in \textit{IEEE Aerospace Conference}, 2002

\bibitem{pyramid} P. Burt, E. Adelson, ``The Laplacian Pyramid as a Compact Image Code," in \textit{IEEE Transactions on Communications}, vol. 31, no. 4, pp. 532-540, 1983

\bibitem{kalman} M. Herbert, C. Caillas, E. Krotkov, I. S. Kweon and T. Kanade, ``Terrain mapping for a roving planetary explorer," in \textit{Proceedings of the International Conference on Robotics and Automation}, pp. 997-1002, 1989

\bibitem{Brownian} M. Müser, W. Dapp, Wolf, R. Bugnicourt, P. Sainsot, N. Lesaffre, T. Lubrecht, B. Persson, K. Harris, A. Bennett, K. Schulze, S. Rohde, P. Ifju, T. Angelini, H. Esfahani, M. Kadkhodaei, S. Akbarzadeh, J. Wu, G. Vorlaufer and J. Greenwood, ``Meeting the Contact-Mechanics Challenge.", in \textit{Tribology Letters}, 2017 

\bibitem{darts} A. Jain, ``DARTS - Multibody Modeling, Simulation and Analysis Software", in \textit{Multibody Dynamics} 2019. Springer International Publishing, p. 433-441, 2019

\end{thebibliography}
\end{document}